%% file: main.tex
\documentclass[conference]{IEEEtran}
\usepackage{cite}
\usepackage{amsmath,amssymb,amsfonts}
\usepackage{algorithmic}
\usepackage{graphicx}
\usepackage{textcomp,comment}
\usepackage{xcolor}
\def\BibTeX{{\rm B\kern-.05em{\sc i\kern-.025em b}\kern-.08em
    T\kern-.1667em\lower.7ex\hbox{E}\kern-.125emX}}

\input{math_commands.tex}

\begin{document}

\title{On the effectiveness of Randomized Signatures as Reservoir for Learning Rough Dynamics
%Randomized Signatures as Reservoir for Learning Rough Dynamics: an Empirical Study %Reservoir Computing using Randomized Signatures: an Empirical Study
}

\author{\IEEEauthorblockN{1\textsuperscript{st} Enea Monzio Compagnoni}\IEEEauthorblockA{\textit{Department of Mathematics} \\
\textit{and Computer Science} \\
\textit{University of Basel}\\
Basel, Switzerland \\enea.monziocompagnoni@unibas.ch}
\and
\IEEEauthorblockN{2\textsuperscript{nd} Anna Scampicchio}
\IEEEauthorblockA{\textit{Institute for Dynamic Systems and Control} \\
\textit{ETH Z\"urich}\\
Z\"urich, Switzerland \\
ascampicc@ethz.ch}
\and
\IEEEauthorblockN{3\textsuperscript{rd} Luca Biggio}
\IEEEauthorblockA{\textit{Department of Computer Science} \\
\textit{ETH Z\"urich}\\
Z\"urich, Switzerland \\
luca.biggio@inf.ethz.ch}
\and
\IEEEauthorblockN{4\textsuperscript{th} Antonio Orvieto}
\IEEEauthorblockA{\textit{Department of Computer Science} \\
\textit{ETH Z\"urich}\\
Z\"urich, Switzerland\\
antonio.orvieto@inf.ethz.ch}
\and
\IEEEauthorblockN{5\textsuperscript{th} Thomas Hofmann}
\IEEEauthorblockA{\textit{Department of Computer Science} \\
\textit{ETH Z\"urich}\\
Z\"urich, Switzerland \\
thomas.hofmann@inf.ethz.ch}
\and
\IEEEauthorblockN{6\textsuperscript{th} Josef Teichmann}
\IEEEauthorblockA{\textit{Department of Mathematics} \\
\textit{ETH Z\"urich}\\
Z\"urich, Switzerland \\
josef.teichmann@math.ethz.ch}
}

\author{\IEEEauthorblockN{Enea Monzio Compagnoni\IEEEauthorrefmark{1},
Anna Scampicchio\IEEEauthorrefmark{2},
Luca Biggio\IEEEauthorrefmark{3},
Antonio Orvieto\IEEEauthorrefmark{3}, \\Thomas Hofmann\IEEEauthorrefmark{3} and
Josef Teichmann\IEEEauthorrefmark{4}}
\IEEEauthorblockA{\IEEEauthorrefmark{1}Department of Mathematics and Computer Science, 
University of Basel, Basel, Switzerland,\\
Email: enea.monziocompagnoni@unibas.ch}
\IEEEauthorblockA{\IEEEauthorrefmark{2}Institute for Dynamic Systems and Control, ETH Z\"urich, Z\"urich, Switzerland.  Email: ascampicc@ethz.ch}
\IEEEauthorblockA{\IEEEauthorrefmark{3}Department of Computer Science, ETH Z\"urich, Z\"urich, Switzerland \\Emails: $\{$luca.biggio, antonio.orvieto, thomas.hofmann$\}$@inf.ethz.ch}
\IEEEauthorblockA{\IEEEauthorrefmark{4}Department of Mathematics, ETH Z\"urich, Z\"urich, Switzerland.  Email: josef.teichmann@math.ethz.ch}
}
\maketitle

\begin{abstract}
Many finance, physics, and engineering phenomena are modeled by continuous-time dynamical systems driven by highly irregular (stochastic) inputs. A powerful tool to perform time series analysis in this context is rooted in rough path theory and leverages the so-called Signature Transform. This algorithm enjoys strong theoretical guarantees but is hard to scale to high-dimensional data. In this paper, we study a recently derived random projection variant called  
 Randomized Signature, obtained using the Johnson-Lindenstrauss Lemma. We provide an in-depth experimental evaluation of the effectiveness of the Randomized Signature approach, in an attempt to showcase the advantages of this reservoir to the community. Specifically, we find that this method is preferable to the truncated Signature approach and alternative deep learning techniques in terms of model complexity, training time, accuracy, robustness, and data hungriness.

%time series analysis (and system identification).continuous-time dynamical systems driven by highly irregular inputs. 
%applications: finance, (physics), biomedical engineering
\end{abstract}


\section{Introduction}
We consider dynamical systems that are described by the stochastic differential equation
\begin{equation}
    dY_t = f(Y_t)dX_t,\qquad Y_0 = y_0 \in \mathbb{R}^m,
    \label{eq:sde}
\end{equation}
where $f(\cdot): \mathbb{R}^m \rightarrow \mathbb
R^d$ is an unknown smooth map, and $X:[0,T]\rightarrow \mathbb{R}^d$ is a piece-wise smooth stochastic process, also known as \textit{control}, forcing the system evolution\footnote{For instance, letting $d=2$, one can have $X_t = [t, \quad W_t]^{\top}$, where $W_t$ is a $1-$dimensional Wiener process.}.
The problem under investigation consists in predicting the solution $\bar{Y}_t$ of \eqref{eq:sde} given a new, unseen control $\bar{X}_t$, using an algorithm trained on a set of control-output trajectories.
Two main challenges %in this problem 
are the fact that data is observed on a possibly unevenly spaced time grid, 
% observations are made on a "possibly irregular" time grid,
and that controls are often highly irregular (we will mainly focus on rough paths, which are formally defined, e.g., in \cite{lyonsqian2002}, Chapter 3). This setting is of particular interest, e.g., in high-frequency trading, as tick-level prices are not observed regularly in time \cite{fu2011review}, and where fundamental quantities such as the volatility of prices are rough \cite{gatheral2018volatility}.\\

A first strategy consists in performing system identification, i.e., learning $f(\cdot)$ from data. The theory is well established for the particular case of linear systems, both for parametric \cite{Ljung} and non-parametric approaches leveraging the theory of Reproducing Kernel Hilbert Spaces \cite{Pillonetto2022, Aronszajn1950TheoryOR}, and is mainly deployed for discrete-time systems; its extension to the continuous-time case is investigated, e.g., in \cite{garnier2014advantagect} and references therein. Nonlinear system identification is an active area of research, and the methods currently deployed revolve around kernel techniques \cite{PILLONETTO2018321}, sparse regression \cite{sindy2016}, random features \cite{NIPS2007_013a006f,rudi2017generalization}, and deep neural networks (see, e.g., \cite{NNARXPaper} for discrete-time models, and \cite{FORGIONE202169} for continuous-time ones). 

The advantage of estimating $f$ is that it allows for retrieving the output trajectory also for different initial conditions and different time intervals. % $[0,T]$. 
However, solving the identification problem of a general continuous-time system is typically hard if observations are made on an uneven time grid and $f$ is nonlinear. Moreover, if the control signal is highly irregular, as typical in finance and physical models \cite{lyons2014rough,kloeden2013numerical}, integrating the differential equation with the estimated $f$ is far from trivial. Thus, an alternative viewpoint consists in focusing on estimating the solution $Y_t$ directly. To this aim, deep neural networks have been successfully deployed \cite{fawaz2019deep, gamboa2017deep}: see also the works on neural controlled differential equations \cite{kidger2020neural,morrill2021neural}. Nevertheless, their outstanding performance comes at the price of over-parametrization, data hungriness, and expensive training cost \cite{neyshabur2018role,marcus2018deep}. Furthermore, %even if sufficient data is available, 
the resulting models learn representations of the input data that are highly specialized to the training task. In addition, the remarkable performance of these methods is often the result of a substantial engineering effort and is not supported by theoretical results.

Another approach consists in reservoir computing \cite{rc}, in which learning is divided into two phases: first, data go through an untrained reservoir that extracts a set of task-independent features; second, a simple and efficient-to-train linear map (the readout map) projects such features into the desired output. An example is Echo State Networks \cite{echo}. The critical point is that the design of the reservoir determines the expressiveness of the features, and several alternatives can be found in the literature (see \cite{Gauthier_2021} and references therein).

A powerful reservoir is offered by the \textit{Signature Transform}, often simply referred to as Signature, stemming from rough path theory~\cite{lyons2010uniqueness,friz2020course}. The Signature of a path is an infinite-dimensional tensor. Intuitively, it consists of enhancing the path with additional curves corresponding to iterated integrals of the curve with itself. A strong mathematical result supports the choice of the Signature as a reservoir: it can be shown \cite{levin2016learning} that the solution of a rough differential equation can be approximated arbitrarily well by a linear map of the Signature of the controls. On the other hand, it is often the case that this reservoir is very high-dimensional, and hence particularly expensive to calculate and computationally intractable for use in downstream tasks. Furthermore, the high dimensionality of the Signature poses additional challenges for modern gradient-based optimizers, as convergence rates suffer from a linear dependence in the model dimension~\cite{bottou2018optimization}.\\
Inspired by the remarkable theoretical properties of the Signature reservoir and motivated to fix its practical pitfalls, the so-called \textit{Randomized Signatures} was introduced in \cite{cuchiero2021discrete,cuchiero2021continuous}. The Randomized Signature is obtained by numerically integrating a set of random linear stochastic differential equations driven by the control path. Importantly, based on a non-trivial Johnson-Lindenstraus argument, \cite{cuchiero2021continuous} showed that calculating the Randomized Signature of a path this way is equivalent to projecting its Signature using a random linear operator. These random features provably retain the expressive power of Signature, yet dramatically reduce the computational burden. In fact, differently from \cite{morrill2020generalised}, calculating the Randomized Signature does not require computing the (truncated) Signature of the path: the projection can be obtained directly in the compressed space. However, the lack of an in-depth experimental study comparing its performance to Signature, Reservoir Computing, and Deep Learning limits its popularity to the theoretical community.

The contribution of the present paper is twofold: first, we extend the theoretical analysis of Randomized Signature, using results from Malliavin Calculus to prove that the Randomized Signature has the power of representing the behavior of any dynamical system of interest; second, we provide a rich set of experiments showing that this approach achieves performances comparable with, if not superior to, competitive Deep Learning, Reservoir Computing, and Signature-based models. In particular, we find that Randomized Signature requires less trainable parameters, i.e. has lower model complexity, which in turn implies a reduced training time and memory usage. In terms of performance, our models are more accurate out-of-sample, more robust, and less data-hungry -- especially in high dimensions.

\paragraph*{Notation} The canonical basis for $\mathbb R^d$ will be denoted as $\{e_{1}, \dots, e_{d}\}$. The symbol $\otimes$ represents a tensor product: e.g., $e_i \otimes e_j$ is the $d\times d$ matrix of all zeros except for the term at the $i-$th row and $j-$th columns, which equals 1. In general, $\left(\mathbb{R}^{d}\right)^{\otimes l}$ is the space of tensors of shape $(d, \ldots, d)$ given by $\mathbb{R}^{d} \otimes \cdots \otimes \mathbb{R}^{d}$ for $l$ times. The tensor algebra on $\mathbb{R}^d$, and its truncated version of order $M \geq 0$, are written as $\mathcal{T}\left(\mathbb{R}^{d}\right):=\prod_{l=0}^{\infty}\left(\mathbb{R}^{d}\right)^{\otimes l}$ and $\mathcal{T}^{M}\left(\mathbb{R}^{d}\right):=\prod_{l=0}^{M}\left(\mathbb{R}^{d}\right)^{\otimes l}$, respectively. Given two vector fields $V_1$ and $V_2$ mapping $\mathbb{R}^k$ into itself, and denoting with $DV_i(z)$ the Fr\'echet derivative of $V_i$ evaluated at $z \in \mathbb{R}^k$, the Lie bracket is defined as $[V_1,V_2](z) = DV_1(z) V_2(z) - DV_2(z) V_1(z)$.

\section{Background}
Referring to the stochastic differential equation \eqref{eq:sde}, we let the control $X = \left( X^{1}, \cdots, X^{d} \right): [0,T] \to \mathbb{R}^{d}$ be a continuous and piece-wise smooth path -- in particular, we will mainly regard $X$ as a rough path\footnote{For the rigorous definition, we refer the reader to, e.g., \cite{ded,lyons2014rough}, due to space limitations. For ease of visualization, one can think of $X$ as a $d-$dimensional fractional Brownian motion with Hurst coefficient $H > 1/4$ (Theorem D.3.2, \cite{Biagini2008}).} on $\mathbb{R}^d$. We start by defining its \textit{Signature}, which is a tensor of iterated integrals of $X$ with itself. 

%Let us now define the Signature for the control $X = \left( X^{1}, \cdots, X^{d} \right)$ introduced in \eqref{eq:sde}, which consists in the tensor algebra of iterated integrals of $X$ with itself. 
\begin{definition}[Signature]
For any $t \in [0,T]$, the Signature of %a continuous piece-wise smooth path 
$X: [0,T] \to \mathbb{R}^{d}$ on $[0,t]$ is the countable collection $
\mathbf{S}_t := \left(1, S_{t}^{1}, S_{t}^{2}, \ldots\right) \in \mathcal{T}\left(\mathbb{R}^{d}\right)
$
where, for each $l \geq 1$, the entries $S_{t}^{l}$ are %the iterated integrals 
defined as
\begin{equation}
\small
    S_{t}^{l}:= \sum_{\substack{\left(i_{1}, \ldots, i_{l}\right) \\\in\{1, \ldots, d\}^{l}}} \left(\int_{0 \leq s_{1} \leq \cdots \leq s_{l} \leq t} d X^{i_{1}}_{s_{1}} \ldots d X^{i_{l}}_{s_{l}}\right) e_{i_{1}} \otimes \dots \otimes e_{i_{l}}.
    \notag
\end{equation}
\end{definition}
Given that this object is infinite-dimensional, to actually compute the reservoir, one can use only a finite amount of terms $S_{t}^{l}$. Therefore, we consider the following object:
\begin{definition}[Truncated Signature]
    The Truncated Signature of $X$ of order $M \geq 0$ is defined as 
\begin{equation}
\small
    \mathbf{S}^{M}_{t} := \left(1, S_{t}^{1}, \ldots,S_{t}^{M} \right) \in \mathcal{T}^{M}\left(\mathbb{R}^{d}\right).
\end{equation}
\end{definition}
To give an intuition on how to compute the (truncated) Signature, we provide the following
\begin{example}
Let $X:[0,T]\to\R$; then $S^{1}_{t} =\int_0^t dX_s$, which is exactly $X_t-X_0$. To get $S^{2}_{t}$, we instead have to compute the following iterated integral: $S^{2}_{t} = \int_0^t\left(\int_0^v dX_s\right) dX_v$. Iterated integrals of higher order $S^{j}_{t}$ are computed in a similar way, by iteratively integrating the path $j$ times. As a practical example, let $X_t = t$. Then it is easy to see that $S^{j}_{t} = \frac{t^j}{j!}$. Now let $Y_t$ be an analytic function of time for which we have $Y_t = \sum_{j=0}^\infty Y^{(j)}_0 \frac{t^j}{j!}$. Taylor's theorem combined with the previous computation implies that $Y$ can be approximated as a linear map of the Truncated Signature of $t$. Finally, note that $S^{j}_{t}$ gets smaller and smaller in magnitude as $j$ increases, which can be proven in general. This suggests that the Truncated Signature can be safely used to approximate $Y$. \hfill$\square$
\end{example}

The following result given in \cite{cuchiero2021continuous} proves that the solutions of differential equations of the type given in \eqref{eq:sde} can be expanded in terms of the iterated integrals stored in the Signature:
\begin{theorem}[Theorem 2.3, \cite{cuchiero2021continuous}]
\label{theorem:SigReservoir}

Let $V_{i}: \mathbb{R}^{m} \rightarrow \mathbb{R}^{m}, i=1, \ldots, d$ be vector fields regular enough such that $d Y_{t}=\sum_{i=1}^{d} V^{i}\left(Y_{t}\right) d X^{i}_{t}, Y_{0} = y \in \mathbb{R}^{m}$, admits a unique solution $Y_{t}: [0,T] \rightarrow \mathbb{R}^{m}$. Then, for any smooth test function $F: \mathbb{R}^{m} \rightarrow \mathbb{R}$ and for every $M \geq 0$ there is a time-homogeneous linear operator $L: \mathcal{T}^{M}\left(\mathbb{R}^{d}\right) \to \mathbb{R}$, which depends only on $\left(V_{1}, \ldots, V_{d}, F, M, y\right)$, such that
\begin{equation}
F\left(Y_{t}\right)=L\left(\mathbf{S}^{M}_{t}\right)+\mathcal{O}\left(t^{M+1}\right),\quad t \in [0,T].
\end{equation}
\end{theorem}

Interpreting the linear operator $L$ as a readout layer, such a result strongly motivates the use of (truncated) Signature as a valuable reservoir under rough dynamics (note that smooth controls constitute a particular case).\\ 
The drawback of using $\mathbf{S}^{M}_{t}$ is that it has an $\mathcal{O}(d^M)$ computational complexity\footnote{Indeed, consider $d=2$: ${S}^{2}_{t}$ is a $2\times 2$ matrix with elements $\int_0^t\left(\int_0^v dX^1_s\right) dX^1_v$, $\int_0^t\left(\int_0^v dX^1_s\right) dX^2_v$, $\int_0^t\left(\int_0^v dX^2_s\right) dX^1_v$ and $\int_0^t\left(\int_0^v dX^2_s\right) dX^2_v$. For $M=3$, the object to compute is instead a $2\times 2\times 2$ tensor, containing all integrals of the type $\int_0^t\left(\int_0^w\left(\int_0^v dX^{i_1}_s\right) dX^{i_2}_v\right)dX^{i_3}_w$ for all $i_1,i_2, i_3\in\{1,2\}$. Hence, the complexity --- as well as the dimensionality of the features --- scales exponentially in $M$.}, which becomes intractable for high-dimensional systems and/or for large values of $M$ aimed at obtaining a finer representation of the solution. %good accuracy level.
To cope with this issue, instead of calculating the Signature, one can extract a new quantity, called Randomized Signature, which is easier to compute and inherits the expressiveness and inductive bias of the Signature. 
\begin{definition}[Randomized Signature]
    Given $k \in \mathbb{N}$ and random matrices $A_{1}, \ldots, A_{d}$ in $\mathbb{R}^{k \times k}$, random shifts $b_{1}, \ldots, b_{d}$ in $\mathbb{R}^{k \times 1}$, random starting point $z$ in $\mathbb{R}^{k}$, and any fixed activation function $\sigma$, the Randomized Signature of $X$ in $t \in [0,T]$ is the solution of the differential equation
\begin{equation}
\label{eq:Rsig}
d Z_{t}=\sum_{i=1}^{d} \sigma\left(A_{i} Z_{t}+b_{i}\right) d X^{i}_{t}, \quad Z_{0}= z \in \mathbb{R}^{k}.
\end{equation}
\end{definition}
The Randomized Signature is constructed in \cite{cuchiero2021continuous} as a random projection of the Truncated Signature according to an argument based on the Johnson-Lindenstrauss Lemma \cite{vempala2005random}. We refer the reader to \cite{cuchiero2021continuous} for all the details on the theoretical derivation. The key message is the following: 
\begin{theorem}[Informal]
    For any number of features $k$ big enough, the Randomized Signature of $X$ defined in \eqref{eq:Rsig} can be linearly mapped to the solution of any differential equation controlled by it, up to a small error vanishing at $k \to\infty$.\label{theorem:RandSig}
\end{theorem} 
This result leads to the practical recipe for extracting the Randomized Signature, summarized in Algorithm \ref{alg::alg1}.
\begin{algorithm}[H]
\caption{Generate Randomized Signature}\label{alg::alg1}
\begin{algorithmic} 
\REQUIRE $X\in\mathbb{R}^d$ sampled at $0=t_{0}< \dots < t_{N}=T$, dimension $k$ of the Randomized Signature, and activation function $\sigma$.
\STATE  {Initialize: } $Z_0 \in \mathbb{R}^{k}, A_i \in \mathbb{R}^{k \times k},  b_i \in \mathbb{R}^{k}$ to have i.i.d. standard normal entries.
%\STATE  \textbf{Initialize: } $\{Z_0\}_j \sim \mathcal{N}(0,1), \{A_i\}_{jl} \sim \mathcal{N}(0,1),  \{b_i\}_{jl} \sim \mathcal{N}(0,1)$ to have iid standard normal entries
\FOR{$n = 1,\cdots, N$}
\STATE \small$Z_{t_{n}} = Z_{t_{n-1}} + \sum_{i=1}^{d} \sigmoid\left(A_{i}Z_{t_{n-1}} + b_{i}\right)\left(X^{i}_{t_{n}}-X^{i}_{t_{n-1}}\right)$
\ENDFOR
\end{algorithmic}
\end{algorithm}
The computational complexity for calculating $Z$ is $\mathcal{O}(k^2d)$, and its dimensionality is $\mathcal{O}(k)$. In Section~\ref{sec:compression} we show experimentally that, in order to match the approximation capabilities of the Truncated Signature of order $M$, the number $k$ of required Randomized Signatures is fairly small -- in particular, it is not exponential in $M$. This confirms that working with Randomized Signatures is often less computationally demanding and results in lower-dimensional -- yet expressive -- features.

\section{Randomized Signature as Reservoir: the procedure} Combining Theorems \ref{theorem:SigReservoir} and \ref{theorem:RandSig}, one can perform linear (ridge) regression to find the sought readout map. This is computed using observed (sampled) control-output trajectories, and can then be used to predict the solution of \eqref{eq:sde} given a new control sequence.
The complete procedure for retrieving the output sequence $\bar{Y}$ given a new control $\bar{X}$ is summarized in Algorithm \ref{alg::alg2}. 
\begin{algorithm}[h!]
\caption{Simulate solution of \eqref{eq:sde}}\label{alg::alg2}
\begin{algorithmic} 
\REQUIRE Time grid $\mathcal{D} = \left\{0=t_{0}, \cdots, t_{N}=T \right\}$; $N_{train}$ input-output trajectories indexed by $m$, $\{(X_t(m),Y_t(m))\}_{t\in \mathcal{D}}$, with common initial condition $y_0 \in \mathbb{R}^m$; new control $\{\bar{X}_t\}_{t \in \mathcal{D}}$; order of Randomized Signature $k$; regularization parameter $\lambda$. 
\FOR{$m=1,...,N_{train}$}
\STATE compute the Randomized Signature $\{Z_t(m)\}_{t \in \mathcal{D}}$ via Algorithm \ref{alg::alg1}.
\ENDFOR
\STATE Define $\mathbf{Y} \in \mathbb{R}^{(N+1)*N_{train}\times m}$ and $\mathbf{Z} \in \mathbb{R}^{(N+1)*N_{train}\times k}$ such that $$\mathbf{Y} = {\scriptsize \begin{bmatrix} Y_{t_0}(1)^{\top}\\ \vdots \\ Y_{t_N}(1)^{\top}\\ \vdots \\ Y_{t_N}(N_{train})^{\top}\end{bmatrix}},\quad \mathbf{Z} = {\scriptsize \begin{bmatrix} Z_{t_0}(1)^{\top}\\ \vdots \\ Z_{t_N}(1)^{\top}\\ \vdots \\ Z_{t_N}(N_{train})^{\top}\end{bmatrix}}$$
\STATE Solve $$ \hat{\beta} = \arg\min_{\beta \in \mathbb{R}^{k\times m}} \|\mathbf{Y} - \mathbf{Z}\beta \|^2 + \lambda\|\beta\|^2.$$
\STATE Compute the Randomized Signature of $\bar{X}$, $\{\bar{Z}_t\}_{t\in \mathcal{D}}$, and store it in $\mathbf{\bar{Z}} = [\bar{Z}_{t_0}, \cdots , \bar{Z}_{t_N}]^{\top}$.
\STATE Retrieve $\mathbf{\bar{Y}} = [\bar{Y}_{t_0}, \cdots, \bar{Y}_{t_N}] = \mathbf{\bar{Z}}\hat{\beta}$.
\end{algorithmic}
\end{algorithm}\\
Note that, while the choice of the activation function $\sigma$ does not affect the theoretical results~\cite{cuchiero2021continuous,cuchiero2021discrete}, selecting it carefully positively impacts expressiveness. Inspired by seminal works on the stability of deep linear networks~\cite{glorot2010understanding} and by the connection to neural ordinary differential equations~\cite{chen2019neural}, it turns out that a good choice for $\sigma$ is a linear function with $\frac{1}{d \times \sqrt{k}}$ as slope\footnote{The dynamics or Randomized Signature is intrinsically exponential. This initialization guarantees that the growth does not depend on the number of controls $d$ nor on the number of features $k$.}. The performance is further affected by the Randomized Signature order $k$, and by the regularization parameter $\lambda$: we select the first via cross-validation, and typically set the latter to the value 0.001. Further investigation into these choices will be carried out in future work.

\section{Theoretical contribution}
We now provide a novel insight into the expressive power of Randomized Signature built as in Algorithm \ref{alg::alg1}. Instead of relying on the theory of rough paths or on approximation estimates, as done in the backbone of Theorem \ref{theorem:RandSig}, we consider tools from Malliavin Calculus \cite{Mal:98,Cass2007DensitiesFR}. 
\begin{theorem}
\label{thm:density}
Let us assume that $ k \geq 2$, that $X$ is a $d$-dimensional Brownian motion, and that the random matrices $A_i \in \mathbb{R}^{k \times k}$ and shifts $b_i \in \mathbb{R}^{k}$ are independent and identically distributed following a law absolutely continuous with respect to the Lebesgue measure on the space of matrices. If the activation function $ \sigma $ is real and analytic, then the Randomized Signature $Z_t$ at time $t$ has a density with respect to Lebesgue measure on $\mathbb{R}^k$ for almost all initial values $Z_0 = z \in \mathbb{R}^{k}$. 
\end{theorem}
\textit{Proof.} 
Considering the vector fields $V_i(z) = \sigma(A_i z + b_i)$ for $i=1,\cdots,d$, it holds that the Lie bracket $[V_i, V_j](z)$ is independent with respect to $V_i(z)$ and $V_j(z)$ almost surely: in fact, independent random samples only meet with probability zero into zero sets of non-constant analytic functions \cite{cuchiero2020deep}. By an inductive argument, it follows that the vector fields $z \to \sigma(A_i z + b_i)$ satisfy H\"ormander condition, i.e. the Lie algebra generated by $\{V_i(z)\}_{i=1}^d$ spans $\mathbb{R}^k$. The conclusion follows by applying, e.g., Theorem 7.4 in \cite{nualart2009malliavin}. \hfill$\blacksquare$\\
Theorem \ref{thm:density} shows that the process $Z_t$ will move in all directions with positive probability: in other words, the obtained coordinate curves $t\to Z_t^i$ form $k$ curves which are almost surely linearly independent in time. As a further consequence, if the control $X_t$ is a Brownian motion, for any partition $\mathcal{D} = \left\{t_{0}, \cdots, t_{N} \right\}$ of $[0,T]$ of size $N+1$, also the sampling $\left( Z^{i}_{t_{0}}, \cdots, Z^{i}_ {t_{N}} \right)$ are almost surely linearly independent among each other if $ k \geq N + 1 $. Therefore, for an appropriate choice of $k \geq N + 1$, Randomized Signature allows representing the behavior of any target dynamical system on time grids $\mathcal{D}$. We highlight that while this result is only proven when the control $X$ is a Brownian motion, Theorem \ref{theorem:RandSig} holds for any, possibly time-varying, rough path (control) X. In the next Section, we show this experimentally.

\section{Numerical Experiments}
We test the effectiveness of the Randomized Signature as a reservoir computer in multiple challenging scenarios. We start by demonstrating the robustness of our approach (Section \ref{sec:randseeds}), as we show that the predictions of Algorithm \ref{alg::alg2} over multiple random initializations are consistent up to a negligible deviation.
Then, we display that it is an effective and efficient low-dimensional compression of the Truncated Signature (Section \ref{sec:compression}), and then we show the resulting advantage in terms of data hungriness and computational time %on a Fractional Ornstein-Uhlenbeck process 
(Section \ref{sec:rsigvstr}). 
 Next, Section \ref{sec:baseline} compares the performance of Randomized Signatures against state-of-the-art techniques for simulation and system identification methods in the presence of a control that is so irregular that it does not even allow a formal definition of Signature. In Section \ref{sec:battery} we deepen such a comparison on the real-world scenario of an electrochemical battery, where measurements are affected by noise. Next, we use the enzyme-substrate model \cite{ingalls2013mathematical} to show the generalization property of the Randomized Signature on out-of-distribution trajectories. Finally, we show on a scalar Langevin equation with double-well potential that our proposed approach can effectively deal with irregularly sampled time grids, which is a main criticality in most of the state-of-the-art methods for trajectory prediction.

\subsection{Robustness over different random initializations}\label{sec:randseeds}

\begin{figure}[h]
\centering
\includegraphics[width=0.65\linewidth]{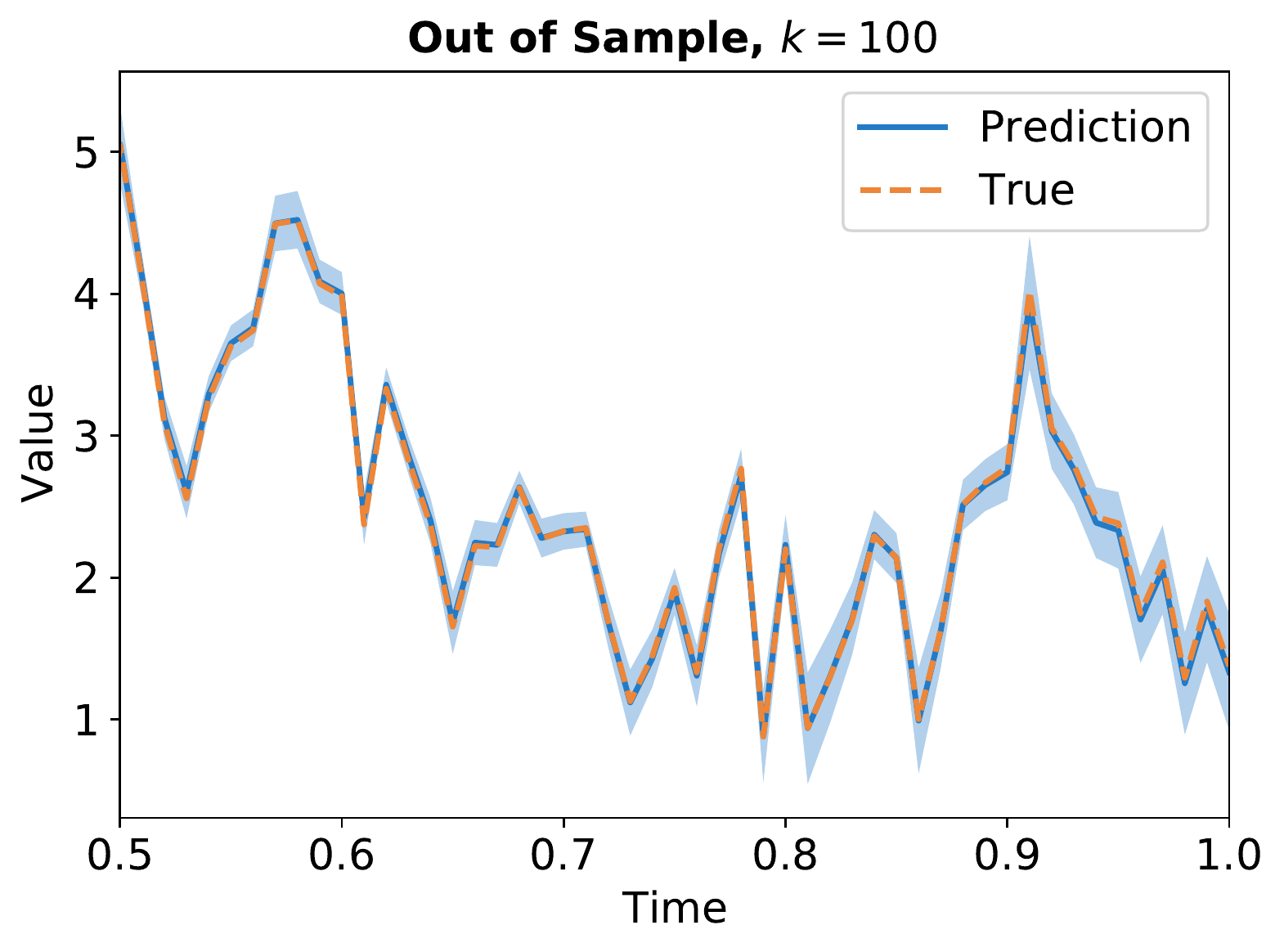}
\caption{Experiment of Section \ref{sec:randseeds}: average prediction with $\pm 3 \times $ standard deviation bounds. }
\label{fig:FOUWN}
\end{figure}

In this experiment, we show that the outputs of Algorithm \ref{alg::alg2} are stable across different realizations of $A_i$, $b_i$, and $Z_0$. We consider the Fractional Ornstein-Uhlenbeck process 
\begin{equation}
\label{eq:frOU}
d Y_{t}=\Theta \left(\mu-Y_{t}\right) d t+\Sigma d B^{(H)}_{t}, \quad Y_{0} = y_{0} \in \mathbb{R}^m,
\end{equation}
where $B_{t}^{(H)}$ is an $m$-dimensional fractional Brownian motion of Hurst parameter $H \in(0,1)$, $\mu \in \mathbb{R}^m$, and $\Theta$, $\Sigma$ are both $m\times m$ positive semi-definite matrices. Relating this model to \eqref{eq:sde}, we have that $X_t = [t,\: (B_t^{(H)})^{\top}]^{\top} \in \mathbb{R}^d$ with $d = m+1$, and $f(Y_t) = [\Theta(\mu - Y_t),\: \Sigma] \in \mathbb{R}^{m\times(m+1)}$. In this experiment, we take $m=1$, $y_{0}=1$ and $\left(\mu=2, \Theta=1, \Sigma=2 \right)$; we let $H=0.2$, and the partition $\mathcal{D}$ of $[0,1]$ is made of $N=101$ equally spaced times. For $10$ different random seeds, we draw different instances of $A_i$, $b_i$, and $Z_0$, generate the reservoir $Z_t$ with $k=100$ and apply Algorithm \ref{alg::alg2} to map $N_{\text{Train}}=100$ train samples of $Z$ into the respective solution $Y_{t}$, to which we add white noise with variance $0.01$. Figure \ref{fig:FOUWN} shows the average prediction ($\pm 3\times$ standard deviation) on a test sample across the above-mentioned $10$ random seeds. 
Because the signal-to-noise ratio is $\approx 50$, this shows that the model is robust to different realizations of the Randomized Signature.

We conclude by highlighting that $k=100$ is a relatively low value with respect to those used in the other experiments: we selected it to make the error bars clearly visible. 
\begin{remark}
We observed the same behavior consistently in all the other proposed experiments, so we omit the Monte Carlo study in the next sections.
\end{remark}

\subsection{Randomized Signature as compression of the Truncated one}\label{sec:compression}
Consider a $10$-dimensional control $X_t=\left[t, (W_{t})^{\top}\right]^{\top}$ where $W_t$ is a $9$-dimensional Brownian motion with independent components, and fix the order of truncation of the Signature to $M=6$. Divide the time interval $[0,1]$ uniformly into $0=t_0 < \cdots < t_N=1$ with $N=100$, and for each element in the grid we compute both the Truncated Signature and the Randomized Signature of order $k$, with $k$ taking values in $\{1, \cdots, 200 \}$. Reshaping the two objects into matrices with dimension $\left(N \times \left(\left(d^{(M+1)}-1\right)/(d-1)-1\right) \right)$ and $N\times k$, respectively, we perform linear regression to find $\mathbf{\beta}\in\mathbb{R}^{k \times \left(\left(d^{(M+1)}-1\right)/(d-1)-1\right)}$ mapping the Randomized Signature into the Truncated Signature. We observed that, in order to obtain an approximation error of $10^{-4}$, we needed the Randomized Signature to be of dimension approximately $k=190$. Therefore, instead of calculating $\left(\left(d^{(M+1)}-1\right)/(d-1)-1\right)=1111110$ integrals per time step, we could just perform $k^{2} d =36100$ calculations per time step, which is 3 times cheaper. 

\subsection{Effectiveness of Randomized versus Truncated Signatures}\label{sec:rsigvstr}

\begin{figure*}
    \centering
    \includegraphics[width=0.28\textwidth]{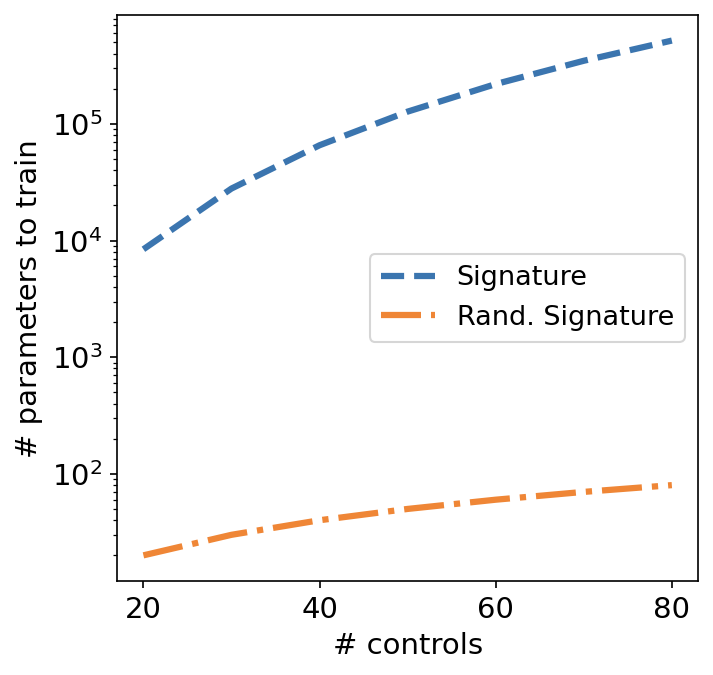}
    \includegraphics[width=0.28\textwidth]{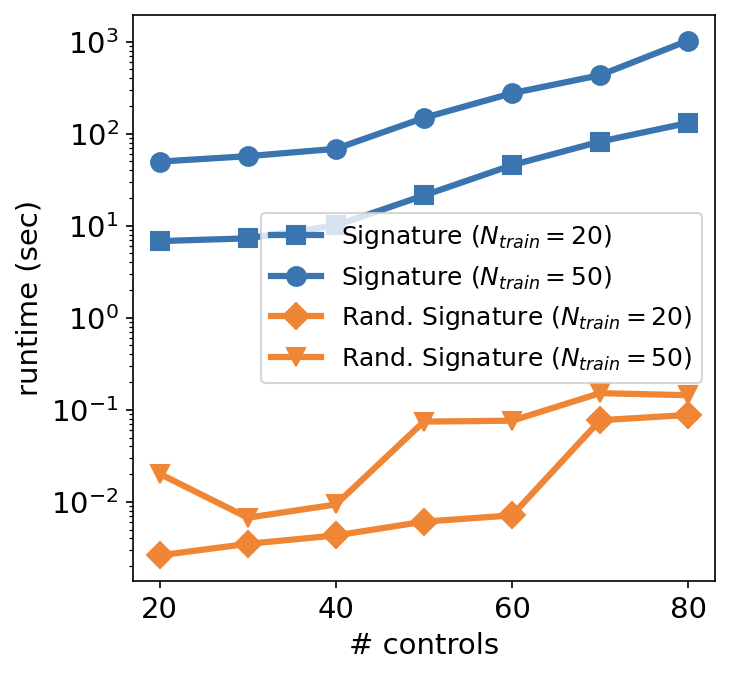}  
    \includegraphics[width=0.28\textwidth]{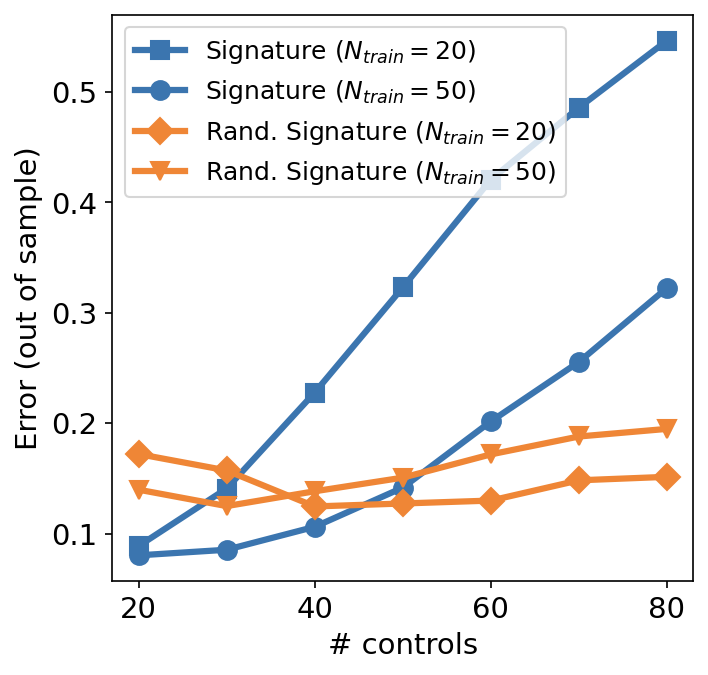}   
    \caption{Experiment of Section \ref{sec:rsigvstr}: Randomized Signature vs. Truncated Signature model. (Left) The number of trainable parameters for Randomized Signature is significantly smaller regardless of the number of controls. (Middle) Truncated Signature is much slower than Randomized Signature in high dimensions. (Right) As opposed to Randomized Signature, the performance of Truncated Signature degrades as the number of controls increases, and even more so when the training set is small, thus indicating data hungriness.}
    \label{fig:comparison_sign}
\end{figure*}

We now deploy Truncated and Randomized Signatures to estimate the dynamics of the Fractional Ornstein-Uhlenbeck process given in \eqref{eq:frOU}.  In this experiment, we fix $y_{0}=\mathbf{1}$, ${\mu}=\mathbf{1}$, ${\Sigma} = I_{d} $, $[\Theta]_{i,j}=i/j$, the partition $\mathcal{D}$ of $[0,1]$ to have $N=101$ equally spaced time steps, and $H=0.3$. 
The order of truncation for the Signature is set to $M=3$, and we consider different experiments with increasing values of $m$ taking range in $\{20, \cdots, 80 \}$. To have a more complete picture, we repeat the experiment in two cases, i.e. when the number of training trajectories is $N_{train} = 20$ and $N_{train} = 50$. To keep the computational cost of extracting features equal to $\mathcal{O}(d^3)$ in both the models given by Truncated and Randomized Signatures, we let $k=d$. As a result, the number of features for the two models are $\mathcal{O}(d^3)$ and $\mathcal{O}(d)$, respectively, strongly impacting the computational time (middle panel of Figure \ref{fig:comparison_sign}). The right panel of Figure \ref{fig:comparison_sign} shows also 
 that the performance of the Truncated Signature degenerates as the underlying optimization problem explodes in dimension, while that of the Randomized one is stable. This result clearly highlights the data hungriness of Signature-based models when the number of dimensions is high.

\subsection{Comparison with baseline methods}\label{sec:baseline} In this experiment, we consider again the Fractional Ornstein-Uhlenbeck process presented in \eqref{eq:frOU} with $m=1$, $y_0=1$, and the same time grid $\mathcal{D}$ of $101$ equally spaced points on [0,1], but we take a Hurst coefficient $H=0.1$ corresponding to a highly irregular control. We benchmark a Randomized Signature of order $k=50$ with the following: (a) Neural Controlled Differential Equations (NCDEs) \cite{kidger2020neural}, a model which parametrizes the vector fields of a latent controlled differential equation of dimension $n_{latent}=100$ with feedforward neural networks with $1$ hidden layer of $n_{nodes}=70$ nodes each, followed by a linear layer mapping the latent variable into the output;  
(b) Echo State Networks (ESN) \cite{echo}, which evolve the input state according to an update rule which is that of an untrained recurrent neural network that is ultimately linearly mapped into the output. We chose the internal state to be of size $50$ (such that we have the same number of trainable parameters as the model based on Randomized Signature) and the activation functions to be hyperbolic tangents. The spectral radius and leaking rate have been selected in cross-validation and set to $0.7$ and $0.4$, respectively; (c) Neural Network Autoregressive model with Exogenous Input (NNARX) as presented in \cite{schoukens2019nonlinear}, i.e. with $n_{a}=n_{b}=12$, $n_{k}=1$, and using a feedforward neural network with input dimension $n_{a}+n_{b}+1=25$ and 2 hidden layers each with $100$ hidden units; (d) a Long Short-Term Memory (LSTM) neural network \cite{lstm} with $2$ hidden recursive layers of dimension $35$. In this experiment, we use $N_{train}=1000$ trajectories to train the models, and $N_{test}=1000$ to test the results. For the NNARX and NCDE models, we minimized the mean square error optimizing with Adam with a learning rate of $0.01$ for $100$ epochs. Similarly, for the LSTM model, we used Adam with a learning rate of $0.001$ for $1000$ epochs. The results, showing the superior performance of the Randomized Signature in terms of accuracy and computational load, are presented in Table \ref{table:FOU_Benchmark}.
\begin{table}[h!]
    \centering 
    \resizebox{\columnwidth}{!}{%
    \begin{tabular}{|c|c|c|c|}
    \hline
         & Average $L^2$ relative error & Training time [s] & \# parameters \\
         \hline
         \scriptsize{RS} & $\mathbf{(1.02 \pm 1.67) \cdot 10^{-5}}$ & $\mathbf{1.59}$ & $\mathbf{50}$ \\
         \scriptsize{NCDE} & $(7.57 \pm 7.95)\cdot 10^{-2}$ & $3296.63$ & $14471$\\
         \scriptsize{ESN} & $(4.24 \pm 3.27)\cdot 10^{-2}$ & $3.01$ & $50$ \\\scriptsize{NNARX} & $(2.96 \pm 8.33)\cdot 10^{-5}$ & $323.73$ & $12801$ \\
         \scriptsize{LSTM} & $(4.49 \pm 6.95) \cdot 10^{-4}$ & $535.21$ & $15436$ \\
         \hline
    \end{tabular}
    }
    \vspace{0.3em}
    \caption{Results for baseline comparison presented in Section \ref{sec:baseline}}
    \label{table:FOU_Benchmark}
\end{table}

\subsection{Real-world experiment: electrochemical battery model with noisy observations}\label{sec:battery}
\begin{figure*}
\centering
\includegraphics[width=0.32\linewidth]{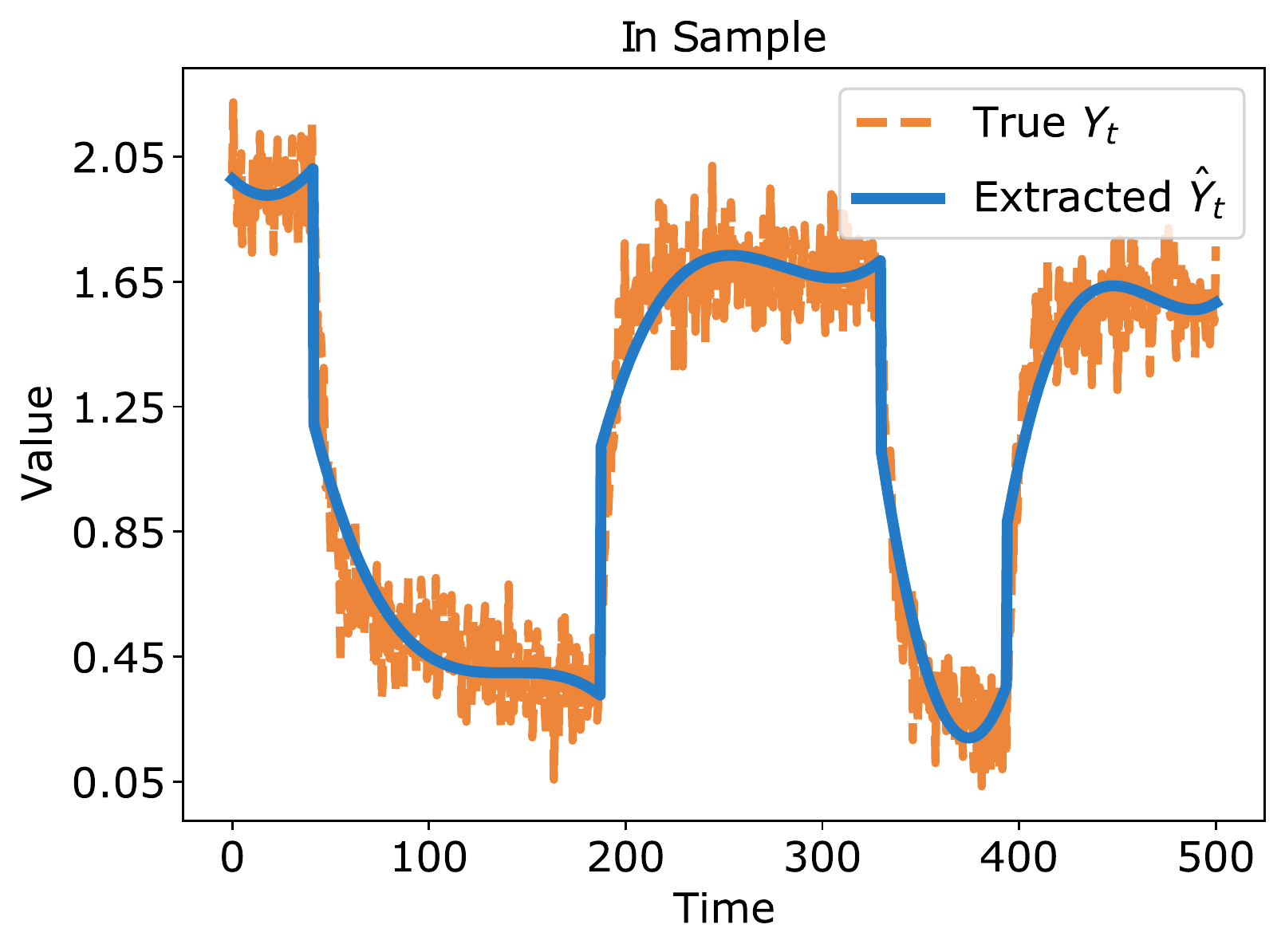}
\includegraphics[width=0.32\linewidth]{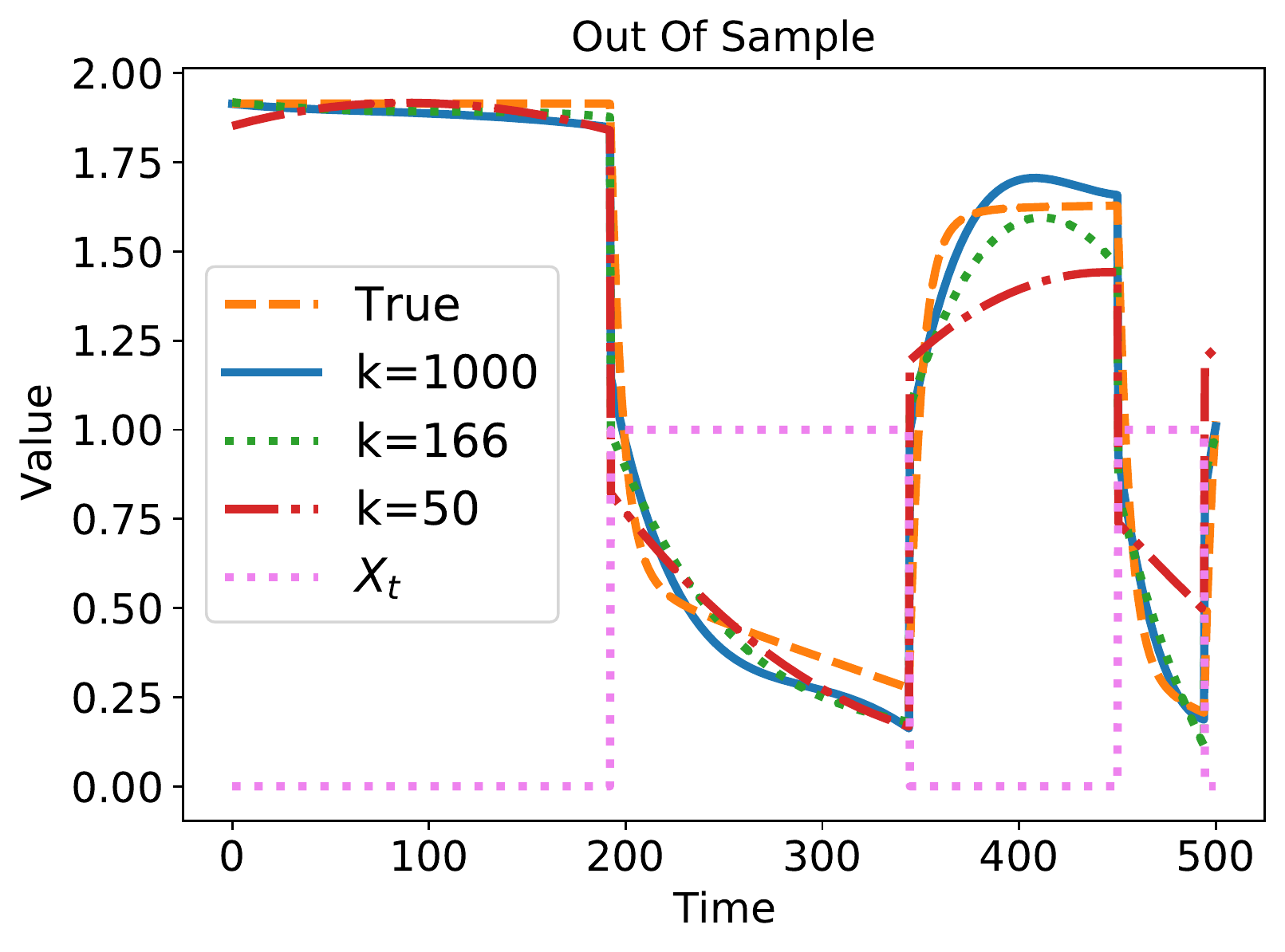}
\includegraphics[width=0.32\linewidth]{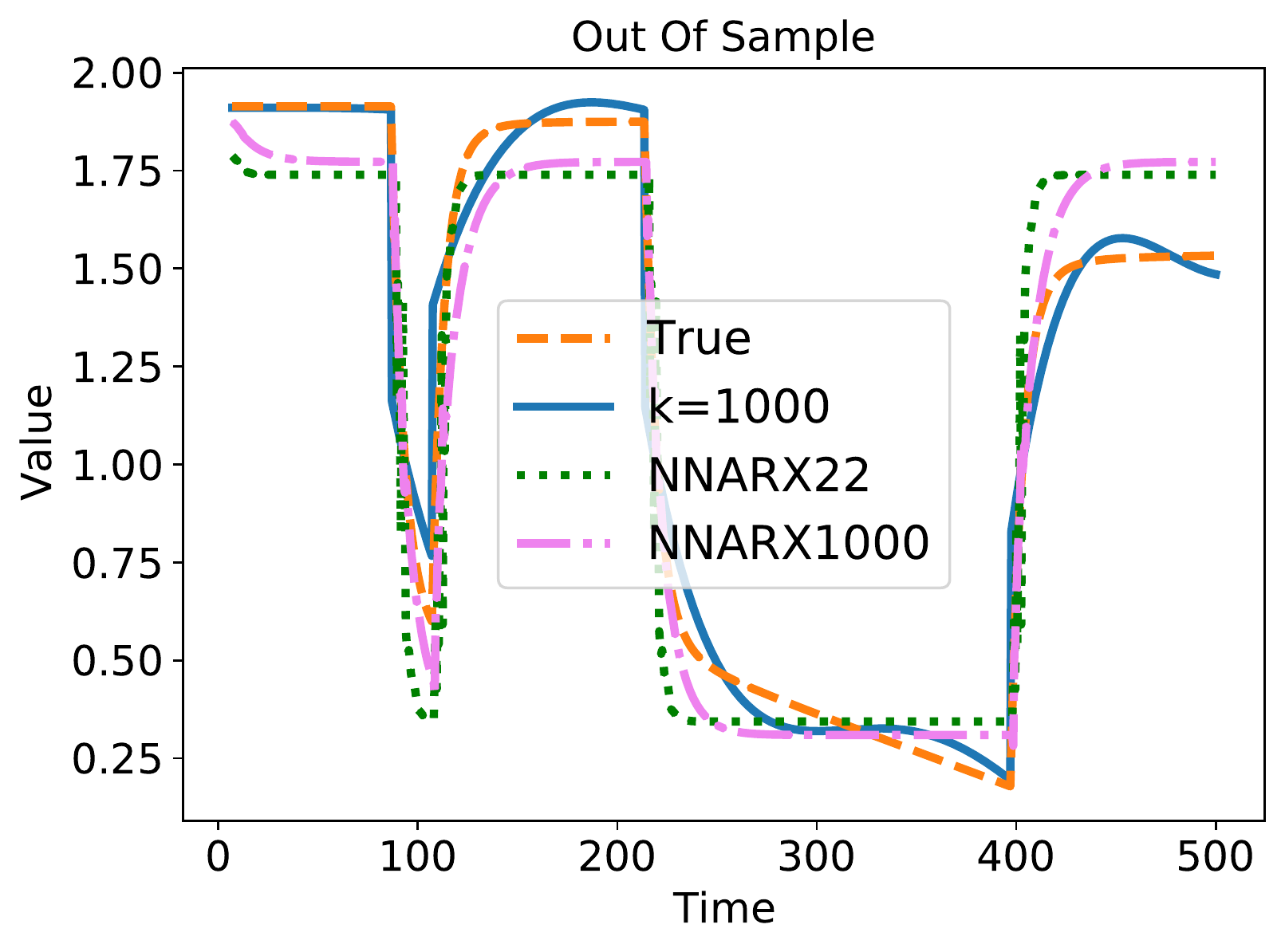}
\caption{\small Experiment of Section \ref{sec:battery} with electrochemical battery model. (Left) Comparison between the ground truth and our prediction on a sample trajectory. (Middle) Predictions on Test Sample for different values of $k$. Consistently, the fit quality improves as $k$ increases. (Right) Comparison with NNARX.}
\vspace{-4mm}
\label{fig:Battery}% 
\end{figure*}

In this experiment, we learn the dynamics of the electrochemical battery model proposed in \cite{daigle2013electrochemistry}, which returns the voltage $Y$ as the current $X$ is injected into the battery. This system is of real-world relevance, as it relies on high-dimensional nonlinear physic-based differential equations that ensure the high fidelity of the simulated data. We use the open-source NASA Prognostic Model Package \cite{2021_nasa_prog_models} to simulate voltage trajectories given input current control paths. On a fixed equally spaced partition $\mathcal{D}$ of $[0,500]$, we model the input current with step functions taking values $0$ or $1$ on random sub-intervals of $[0,500]$. We apply Algorithm \ref{alg::alg2} to map $N_{Train}=1000$ instances of $k$-dimensional Randomized Signature of the controls into the respective solutions to which we add white noise with variance 0.01. We consider $k=\{50, 166, 1000\}$. We compare our results with NNARX -- which, consistently with the experiment in Section \ref{sec:baseline}, is the best-performing benchmark on this task. Specifically, we choose the parameters as $n_{a}=n_{b}=12$ and $n_{k}=1$, which leads to the best results, and we use a feedforward neural network with input dimension $n_{a}+n_{b}+1=25$ and 2 hidden layers each with either $22$ (NNARX22) or $1000$ (NNARX1000) hidden units. We minimized the mean square error optimizing with Adam with a learning rate of $0.01$ for $100$ epochs. The results are presented in Figure \ref{fig:Battery}. Furthermore, Table \ref{table:benchNNARX} presents the out-of-sample comparison in terms of Mean Squared Error with respect to the ground truth averaged across $N_{test} = 1000$ test trajectories. Note that the NNARX22 has around $1000$ trainable parameters just like our model with $k=1000$ and that NNARX1000, which matches our best model in terms of MSE, has around $10^{3}$ more trainable parameters.
\begin{table}[h!]
\centering
\resizebox{\columnwidth}{!}{%
 \begin{tabular}{||c||c c c c c||} 
 \hline
& $k=50$ & $k=166$ & $k=1000$ & NNARX 22 & NNARX 1000 \\ [0.1ex] 
 \hline
$N_{Test}=1000$ & $3.15 \cdot 10^{-4}$ & $2.66 \cdot 10^{-4}$ & $\mathbf{2.59 \cdot 10^{-4}}$ & $3.88 \cdot 10^{-4}$ & $2.62 \cdot 10^{-4}$ \\ [0.1ex] 
 \hline
\end{tabular}
}
\vspace{0.1em}
\caption{Electrochemical battery model, experiment in Section \ref{sec:battery}: MSE Error Comparison}
\label{table:benchNNARX}
\end{table}

\subsection{Out-of-sample generalization on enzyme-substrate model}\label{sec:enzymeout} 
\begin{figure}[H]
\centering
\includegraphics[width=0.72\linewidth]{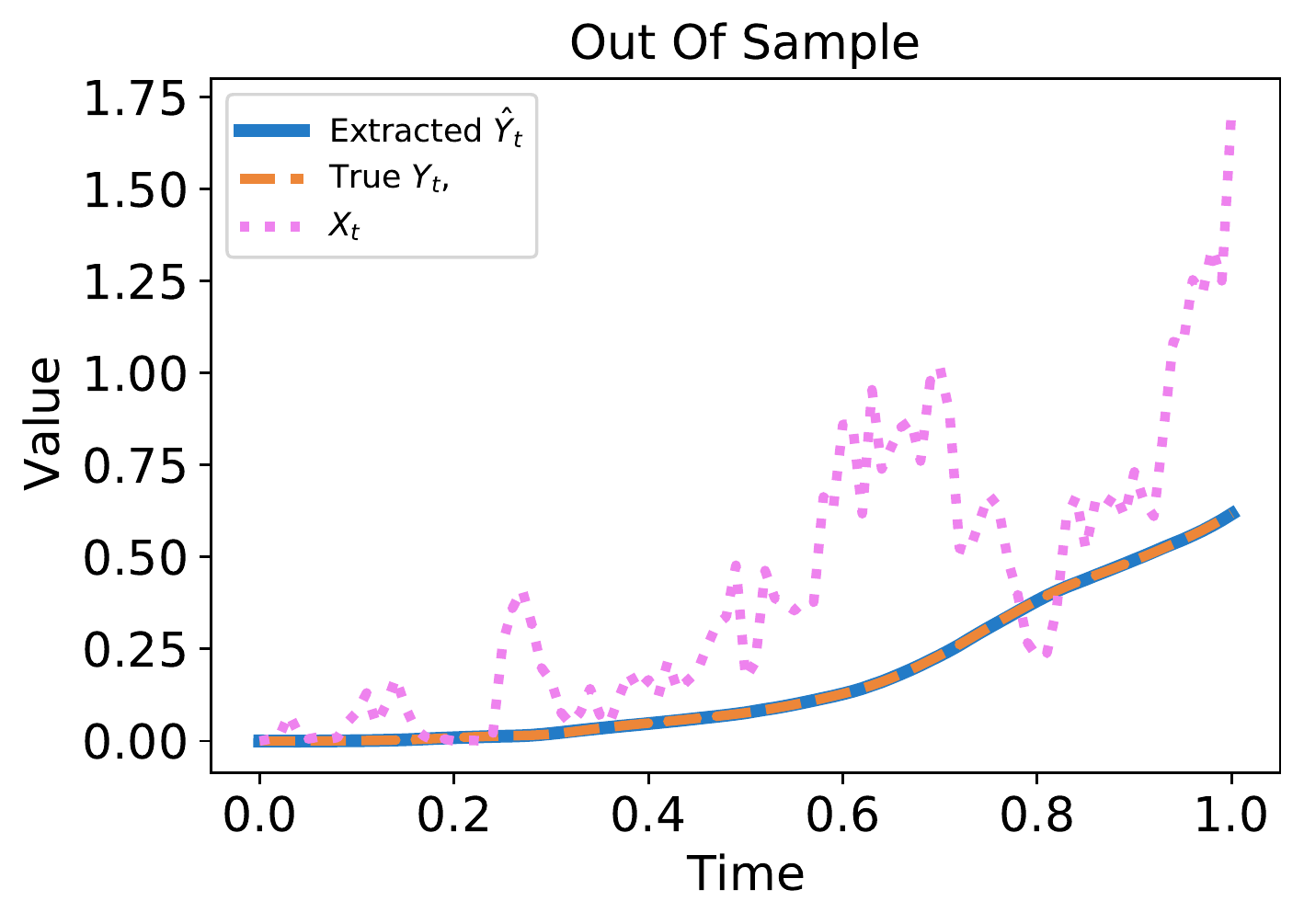}\\
\hspace{0.3em} \includegraphics[width=0.7\linewidth]{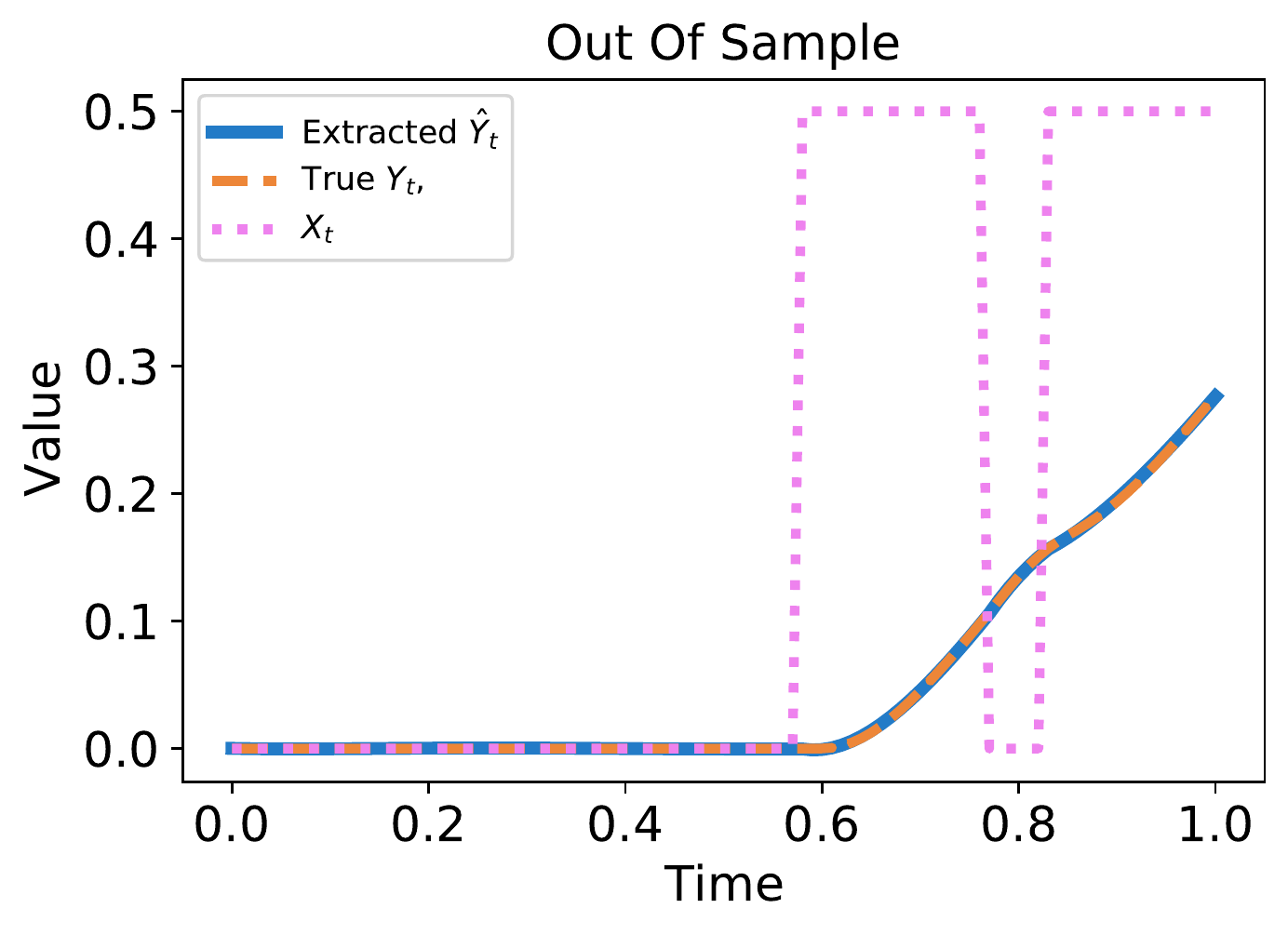}
\vspace{-2mm}
\caption{Experiment of Section \ref{sec:enzymeout}. Enzyme-Substrate Reactions stimulated with: (Top) squared Brownian motion; (Bottom) step function, which is out-of-distribution.}
\label{fig:Enzyme_Comparison}% 
\end{figure}

We consider the controlled differential equation describing the reaction between concentrations of a substrate $S_t$ and of an enzyme $E_t = 1 - S_t$, yielding the enzyme-substrate complex $C_t$ according to the Michaelis-Menten model. Additional substrate is injected through a control $X_t$, and the observed quantity of interest $Y_t$ is the chemical product of the reaction -- for instance, the latter can be glucose obtained from lactose-lactase reaction. The overall kinetics can be described by the model %We consider the controlled differential equations modeling the interaction between a catalyst (in this case, an enzyme) with abundance $C_t$ and a substrate with abundance $S_t$. The substrate~(e.g. lactose) is injected via the control variable $X_t$, it reacts with the enzyme~(e.g.~lactase) and the concentration $Y_t$ of the product chemical~(e.g. glucose) is the observed quantity of interest. The resulting model reads as
\begin{equation*}
\begin{cases}
dS_{t} & = \left(k_{-1}C_{t} - k_1 S_{t}\left(1-C_{t}\right)\right)dt + X_{t}dt\\
dC_{t} & = - \left(k_{-1}C_{t} - k_1 S_{t}\left(1-C_{t}\right)\right)dt - k_2C_{t}dt\\
dY_{t} & = k_2 C_{t}dt.
\end{cases}
\end{equation*}
Following \cite{ingalls2013mathematical}, we choose $(k_1,k_{-1}, k_2) = (30, 1, 10)$, set $\left(S_{0},C_{0},Y_{0}\right) = (0,0,0)$ and consider the evolution on $t \in [0,1]$. We fix the time grid to have $N=101$ equally spaced time steps and the control $X_{t}$ to follow the law of $W^{2}_{t}$ where $W_{t}$ is a $1$-dimensional Brownian Motion~(to ensure positivity). We apply Algorithm \ref{alg::alg2} to map $10^5$ instances of $222$-dimensional Randomized Signature $Z$ of the controls into the respective solution $Y_{t}$.
On the top of Figure \ref{fig:Enzyme_Comparison}, we plot the comparison of the true and the generated time series on a test sample. As we can see, the model has learned to correctly map a trajectory of $X_{t}$ to the respective system response $Y_{t}$. More surprisingly, the bottom of such a figure shows that our model is able to predict the correct output even if we stimulate the system with a substrate injection that follows a completely different law with respect to those used in training, i.e. $X_t=0.5 \cdot \mathbb{1}_{\left\{W^{2}_{t}>0.5\right\}}$. This suggests that the system was correctly identified even out-of-distribution.

\subsection{Test on irregularly sampled grid}\label{sec:irregulargrid}
\begin{figure}[h!]
\centering
\includegraphics[width=0.68\linewidth]{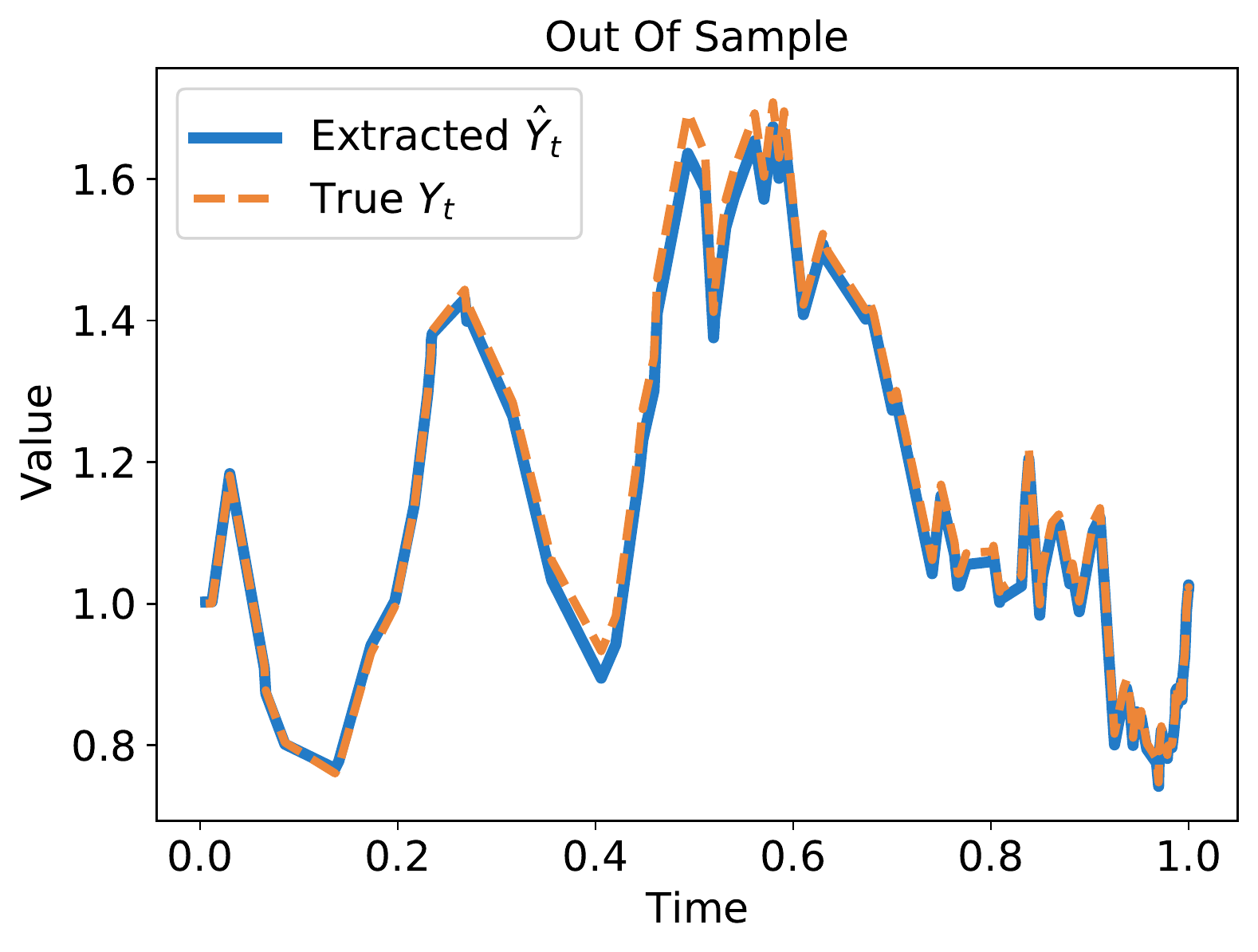}
\vspace{-2mm}
\caption{Experiment of Section \ref{sec:irregulargrid} with Langevin equation. Simulation of an out-of-sample trajectory using an irregularly sampled time grid.}
\label{fig:DW_Irreg_FGMB}% 
\end{figure}

We consider the $1$-dimensional Langevin equation with double-well potential given by
\begin{equation}
\label{equation:1DW}
d Y_{t}=\theta Y_{t} \left(\mu-Y^{2}_{t}\right) d t+\sigma d W_{t}, \quad Y_{0} = y_{0} \in \mathbb{R},
\end{equation}
where $t \in [0,1]$, $W_{t}$ is a $1$-dimensional Brownian motion, and $\left(\mu, \theta, \sigma \right) \in \mathbb{R} \times \mathbb{R}^{+} \times \mathbb{R}^{+}$; in this experiment, we fix $y_{0}=1$ and $\left(\mu=2, \theta=1, \sigma=1 \right)$. For each train and test sample, the partition $\mathcal{D}$ of $[0,1]$ is made of $N$ randomly drawn times. More precisely, $\mathcal{D} = \{0, t_{1}, \cdots, t_{N-1}, 1 \}$ such that $t_{k} = 1/(1-\exp(-s_{k}))$ and $ \{s_{1}, \cdots, s_{N-1} \}$ are $N-2$ independent realizations of a uniform distribution $\mathcal{U}[0,1]$ sorted in increasing order. As a result, the probability that two samples share the same $\mathcal{D}$ is null. We apply Algorithm \ref{alg::alg2} with $N_{\text{Train}}=10000$ train samples, and Figure \ref{fig:DW_Irreg_FGMB} shows the comparison on an out-of-sample generated and true trajectory. Finally, Table \ref{table:DW1_Relat_Error_Irreg} shows the Relative $L^{2}$ Error on $10000$ test samples as we vary the number of time steps $N$ and $k$, and we compare it to the respective experiment in case the time grid is regularly spaced. As we can see, even though the performance is worse than the regularly sampled setup, this technique proves to be anyway reliable on irregularly sampled regimes. 

\begin{table}[ht!]
\centering
\resizebox{\columnwidth}{!}{%
 \begin{tabular}{||c|| c c c||} 
 \hline
& $(N,k) = (11, 111)$ & $(N,k) = (101, 222)$ & $(N,k) = (1001, 332)$\\ [0.1ex] 
 \hline
Irregular & $0.082735$ & $0.016885$ & $0.010902$\\ [0.1ex] 
 \hline
Regular & $0.026759$ & $0.004465$ & $0.003004$\\ [0.1ex] 
 \hline
\end{tabular}
}
\vspace{0.3em}
\caption{\small Relative $L^{2}$ Error comparison with regularly and irregularly sampled grid (Section \ref{sec:irregulargrid})}
\label{table:DW1_Relat_Error_Irreg}
\end{table}

\section{Conclusions}
A challenging problem emerging in a plethora of fields consists in solving a controlled stochastic differential equation. The main difficulties that can arise in this situation may be: (a) the law governing the differential equation is unknown, so one needs to rely on sampled input/output trajectories; (b) the samples are observed on an irregular time grid; (c) the input trajectory is highly irregular, e.g., is a rough path. To cope with them, this work investigated the power of Randomized Signature as a reservoir. Such an approach proved to be very effective in estimating the solution of the stochastic differential equation driven by a new control input, showing its low data hungriness and robustness compared with state-of-the-art system identification and deep learning-based methods. Further investigations will aim at providing deeper theoretical results on the generalization capability of Randomized Signature.

\bibliography{main.bib}
\bibliographystyle{plain}

\end{document}

%% file: math_commands.tex
\usepackage{amsmath,amsfonts,bm}

\def\1{\bm{1}}

\DeclareMathAlphabet{\mathsfit}{\encodingdefault}{\sfdefault}{m}{sl}
\SetMathAlphabet{\mathsfit}{bold}{\encodingdefault}{\sfdefault}{bx}{n}

\newcommand{\R}{\mathbb{R}}

\newcommand{\sigmoid}{\sigma}

\newcommand{\enea}[1]{{\color{blue} Enea: ``#1''}}

\usepackage{tikz}

\usepackage{amsmath}
\usepackage{amssymb}
\usepackage{mathtools}
\usepackage{amsthm}
\usepackage{mathbbol}
\theoremstyle{plain}
\newtheorem{theorem}{Theorem}[section]

\theoremstyle{definition}
\newtheorem{definition}[theorem]{Definition}

\newtheorem{example}[theorem]{Example}
\theoremstyle{remark}
\newtheorem{remark}[theorem]{Remark}

\usepackage{algorithm}
\usepackage{algorithmic}